\documentclass{article}
\usepackage{ijcai23}

\usepackage{times}
\usepackage{soul}
\usepackage{url}
\usepackage[utf8]{inputenc}
\usepackage[small]{caption}
\usepackage{graphicx}
\usepackage{amsmath}
\usepackage{amsthm}
\usepackage{booktabs}
\usepackage{algorithm}
\usepackage{algorithmic}
\usepackage[switch]{lineno}
\usepackage[shortlabels]{enumitem}

\usepackage[linktoc=page,colorlinks=true,linkcolor=RedOrange,citecolor=NavyBlue]{hyperref}
\usepackage[dvipsnames]{xcolor}
\usepackage[capitalise,nosort]{cleveref} % \usepackage{cleveref}
\numberwithin{equation}{section}
\usepackage{amsmath,amsfonts,amssymb,amsthm,bm} % amsmath suite
\usepackage{nicefrac}       % compact symbols for 1/2, etc.
\usepackage{microtype}
\usepackage[super]{nth}     % generate English ordinal numbers
\usepackage{graphicx}       % enhanced support for graphics
\usepackage{crossreftools}  % expandable extraction of cleveref data
\usepackage[capitalise,nosort]{cleveref}      % intelligent cross-referencing

\usepackage{caption}        % show caption on top of the table
\usepackage{subfigure}
\usepackage{booktabs} % for professional tables
\usepackage{multirow}

\usepackage{mathtools,mathrsfs,grffile}
\newtheorem{thm}{Theorem}

\newtheorem{lem}{Lemma}
\newtheorem{prop}{Proposition}

\newtheorem{remark}{Remark}

\newtheorem{assumption}{Assumption}

\theoremstyle{definition}
\newtheorem{defn}{Definition}
\theoremstyle{remark}

\newcommand{\RR}{\mathbb R}

\newcommand{\XX}{\mathbf X}
\newcommand{\bz}{\mathbf{z}}
\newcommand{\bu}{\mathbf{u}}
\newcommand{\bU}{\mathbf{U}}
\newcommand{\bD}{\mathbf{D}}
\newcommand{\bx}{\mathbf{x}}
\newcommand{\bLambda}{\boldsymbol{\Lambda}}

\newcommand{\bA}{\mathbf{A}}

\newcommand{\bL}{\mathbf{L}}

\newcommand{\bH}{\mathbf H}
\newcommand{\bW}{\mathbf W}
\newcommand{\bw}{\mathbf{w}}
\newcommand{\bI}{\mathbf{I}}
\newcommand{\bv}{\mathbf{v}}
\newcommand{\bg}{\mathbf{g}}

\newcommand{\bzero}{\mathbf{0}}

\usepackage[page]{appendix}
\renewcommand{\appendixtocname}{List of Appendices}

\makeatletter
\let\oldappendix\appendices

\g@addto@macro\tableofcontents{%
  \let\tf@toc@orig\tf@toc
}
\renewcommand{\appendices}{%
  \clearpage
  \renewcommand{\thesection}{\Roman{section}}
  \let\tf@toc\tf@toca
  \addtocontents{toca}{\protect\setcounter{tocdepth}{3}}
  \immediate\write\@auxout{%
    \string\let\string\tf@toc\string\tf@toca
  }
  \oldappendix
}%
\g@addto@macro\endappendices{%
  \let\tf@toc\tf@toc@orig
  \immediate\write\@auxout{%
    \string\let\string\tf@toc\string\tf@toc@orig
  }%
}  

\newcommand{\listofappendices}{%
  \begingroup
  \renewcommand{\contentsname}{\appendixtocname}
  \let\@oldstarttoc\@starttoc
  \def\@starttoc##1{\@oldstarttoc{toca}}
  \tableofcontents% Reusing the code for \tableofcontents with different \contentsname and different file handle app
  \endgroup
}

\makeatother

\title{Stability and Generalization of $\ell_p$-Regularized Stochastic Learning for GCN}

\author{
    Anonymous Author(s)
}

\author{
Shiyu Liu$^1$\and
Linsen Wei$^{2}$\footnotemark[1] \and
Shaogao Lv$^3$\footnotemark[2] \And
Ming Li$^4$\footnotemark[1] 
\affiliations
$^1$University of Electronic Science and Technology of China, China\\
$^2$School of Astronautics, Northwestern Polytechnical University, China\\
$^3$Department of Statistics and Data Science, Nanjing Audit University, China\\
$^4$Key Laboratory of Intelligent Education Technology and Application of Zhejiang Province, Zhejiang Normal University, China
}
\begin{document}

\maketitle
\renewcommand{\thefootnote}{\fnsymbol{footnote}}
\footnotetext[2]{Corresponding author. E-mail: \textsc{lvsg716@nau.edu.cn}}
\footnotetext[1]{Equal contribution.}
\renewcommand{\thefootnote}{\arabic{footnote}}

\begin{abstract}
    Graph convolutional networks (GCN) are viewed as one of the most popular representations among the variants of graph neural networks over graph data and have shown powerful performance in empirical experiments. That $\ell_2$-based graph smoothing enforces the global smoothness of GCN, while (soft) $\ell_1$-based sparse graph learning tends to promote signal sparsity to trade for discontinuity. This paper aims to quantify the trade-off of GCN between smoothness and sparsity, with the help of a general $\ell_p$-regularized $(1<p\leq 2)$ stochastic learning proposed within. While stability-based generalization analyses have been given in prior work for a second derivative objectiveness function, our $\ell_p$-regularized learning scheme does not satisfy such a smooth condition. To tackle this issue, we propose a novel SGD proximal algorithm for GCNs with an inexact operator. 
    For a single-layer GCN, we establish an explicit theoretical understanding of GCN with the $\ell_p$-regularized stochastic learning by analyzing the stability of our SGD proximal algorithm. %While such analysis has been performed in recent years for other models with regular data, the massage passing nature embedded in a GCN poses additional challenges in understanding the nature of the stochastic gradient for GCN. We attempt to overcome this by proving that the uniform stability of our GCN depends on the largest absolute eigenvalue of its graph filter, and there exists a stability-sparsity tradeoff with varying $p$. 
    % Several empirical experiments are implemented to validate our theoretical findings.
    We conduct multiple empirical experiments to validate our theoretical findings.	
\end{abstract}

\section{Introduction}
Graph Neural Networks (GNNs) have emerged as a family of powerful model designs for improving the performance of neural network models on graph-structured data.
GNNs have delivered remarkable empirical performance from a diverse set of domains, such as social networks, knowledge graphs, and biological networks \cite{duvenaud2015convolutional,battaglia2016interaction,defferrard2016convolutional,jin2018junction,barrett2018measuring,yun2019graph,zhang2020efficient}. In fact, GNNs can be viewed as natural extensions of conventional machine learning for any data where the available structure is given by pairwise relationships.
%As a popular GNN architecture, Graph Convolutional Network 

The architecture designs of various GNNs have been motivated mainly by spectral domain \cite{defferrard2016convolutional,kipf2016semi} and spatial domain \cite{hamilton2017inductive,gilmer2017neural}. Some popular variants of graph neural networks include Graph Convolutional Network (GCN) \cite{bruna2013spectral}, GraphSAGE \cite{hamilton2017inductive}, Graph Attention Network \cite{velivckovic2017graph}, Graph Isomorphism Network \cite{xu2018powerful}, and among others. 

Specifically, inherited excellent performances of traditional convolutional neural networks in processing image and time series, 
	a standard GCN \cite{kipf2016semi} also consists of a cascade of layers, but operates directly on a graph and induces embedding vectors of nodes based on properties of their neighborhoods. Formally, GCN is defined as the problem of learning filter parameters in the graph Fourier transform. GCNs have shown superior performances on real datasets from various domains, such as node labeling on social networks \cite{kipf2016variational}, link prediction in knowledge graphs
	\cite{schlichtkrull2018modeling}, and molecular graph classification in quantum chemistry \cite{gilmer2017neural}.

Notably, a recent study \cite{ma2021unified} has proven that GCN, even for general messaging passing models, intrinsically performs the $\ell_2$-based graph smoothing signal, which enforces smoothness globally, and the level and smoothness are often shared across the whole graph. As opposed to $\ell_2$-based graph smoothing, $\ell_1$-based methods tend to penalize large values less and thus preserve discontinuity of non-smooth signal better \cite{nie2011unsupervised,wang2015trend,liu2021elastic}. Essentially, $\ell_1$-based methods are equivalent to soft-thresholding operations for iterative estimators and guarantee statistical properties (e.g., model selection consistency). Owning to these advantages, trend filtering \cite{tibshirani2014adaptive}, and graph trend filter \cite{wang2015trend,verma2019stability} indicate that $\ell_1$-based graph smoothing can adapt to the inhomogeneous level of smoothness of signals and yield estimator with $k$-th piecewise polynomial functions, such as piecewise constant, linear and quadratic functions, depending on the order of the graph difference operator.

To enhance the local smoothness adaptivity of GCNs, a family of elastic-type GCNs with a combination of $\ell_2$ and $\ell_1$-based penalties are proposed by \cite{liu2021elastic}, which demonstrate that the elastic GCNs obtain better adaptivity on benchmark datasets and are significantly robust to graph adversarial attacks. 

Under the regularized learning framework, this paper further studies a class of $\ell_p$-regularized learning approaches $(1<p\leq 2)$, in order to trade-off the local smoothness of GCNs and sparsity between nodes. To be precise, it has been shown that an extreme case for $\ell_p$-regularized learning with $p\rightarrow 1$ tends to generate soft-sparsity of solutions \cite{koltchinskii2009sparsity}. In analogy to elastic-type GCNs, general $\ell_p$-regularization can be interpreted as an interpolation of $\ell_1$-regularization and $\ell_2$-regularization. However, in contrast with the elastic-type GCNs, $\ell_p$-regularized GCNs only involve a regularized parameter but leads to some additional technical difficulties in optimization and theoretical analysis. 

The generalization performance of an algorithm has always been a central issue in learning theory, and particularly the generalization guarantees of GNNs has attracted a considerable amount of attention in recent years \cite{verma2019stability,garg2020generalization,liao2020pac,oono2020optimization}.
 
It is worth noting that, although $\ell_1$-regularization possesses a number of attractive properties such as ``automatic" variable selection, the objectiveness with $p=1$ is not a strictly convex smooth function, which has been proved not to be uniform stability \cite{xu2011sparse}. On the other hand, it is known that for $p > 1$, the penalty is a strictly convex function over bounded domains and thus enjoys robustness to some extent. As a bridge of sparse $\ell_1$-regularization and dense $\ell_2$-regularization, $\ell_p$-based learning allows us to explicitly observe a trade-off between sparsity and smoothness of the estimated learners. 

Although this idea is natural within the framework of regularized learning, it still faces many technical challenges when applied to GCNs. First, to address the problem of uniform stability of an SGD algorithm that can induce generalization performance \cite{bousquet2002stability}, existing related analysis for SGD \cite{verma2019stability,hardt2016train} required that the first derivative of the objectiveness is Lipschitz continuous, which is not applicable to $\ell_p$-based GCN. Second, to derive an interpretable generalization bound, it is important to know how such a result depends on the structure of the graph filter and the regularized parameter $p$, as well as the network size and the sample size. 

Overall, our principal contributions can be summarized as follows: 
\begin{itemize}[leftmargin=*]
	\item We introduce $\ell_p$-regularized learning approaches for one-layer GCN to provide an explicit theoretical characterization of the trade-off between local smoothness and sparsity of the SGD algorithm for GCN. Crucially, we analyze how this trade-off between the graph structure and the regularization parameter $p$ affects the generalization capacity of our SGD method.
	\item We propose a novel regularized stochastic algorithm for GCN, i.e., {\it Inexact Proximal SGD}, by integrating the standard SGD projection and the proximal operator. For our proposed method, we derive interpretable generalization bounds in terms of the graph structure, the regularized parameter $p$, the network width, and the sample size. 
	\item To our knowledge, we are the first to analyze the generalization performance of $\ell_p$-SGD in GNNs, which is quite different from existing stability-based generalization analysis for a second derivative objectiveness \cite{verma2019stability,hardt2016train}. We also have to overcome additional challenges in understanding the nature of the stochastic gradient for GCN, posed by the message passing nature in GCN. 
	We conduct several numerical experiments to illustrate the superiority of our method to traditional smooth-based GCN, and we also observe some sparse solutions through our experiments as $p$ is sufficiently close to $1$.

\end{itemize}

%GCNN, containing features information and edge information over graph, have shown superior performance on various domains such as molecular structures, knowledge graphs, biological networks and social networks. 
%Despite their success in practice, a complete theoretical characterization of GCNN models on generalization performance is still lacking.

\subsection{Additional Related Work}
In this subsection, we briefly review two kinds of related works: Generalization Analysis for GNNs and Regularized Schemes on GNNs.

\paragraph{Generalization Analysis for GNNs.}
Some previous work has attempted to address the generalization guarantees of GNNs, including \cite{verma2019stability,garg2020generalization,liao2020pac,oono2020optimization}. However, most of these works established uniform convergence results using classical Rademacher complexity\cite{garg2020generalization,oono2020optimization} and PAC bounds \cite{liao2020pac}. Compared to these abstract capacity notations, the stability concept, used recently for GNN in \cite{verma2019stability}, is more intuitive and directly defined over a specific algorithm. Although the stability for generalization of GNNs has been considered recently by \cite{verma2019stability}, a major difference from their work is that we focus on the trade-off between soft sparsity and generalization of the $\ell_p$-based GCN with varying $p\in(1,2]$. Moreover, we propose a new SGD algorithm for GCN, under which we provide novel theoretical proof for the stability bound of our SGD.

\paragraph{Regularized Schemes on GNNs.}
Regularization methods are frequently used in machine learning, especially the prior literature \cite{wibisono2009sufficient} has studied the influence of $\ell_p$-regularized learning on generalization ability. Note that the previous work only considered the impact of general regularization estimates on stability without a specific algorithm. In addition, their research is based on regular data and does not involve any graph structure. As far as we know, this paper is the first time to consider $\ell_p$-regularized learning in a graph model. 

\section{Preliminaries and Methodology}
In this section, we first describe basic notation on graph and standard versions of GCN. Then we introduce the structural empirical risk minimization for a single-layer GCN model under i.i.d. sampling process, and thus our $\ell_p$-regularized approaches is naturally formulated to estimate the graph filter parameters.

 A graph is represented as $G=(V,E)$, where
$V=\{\nu_1,\nu_2,...,\nu_n\}$ is a set of $n=|V|$ nodes and
$E$ is a set of $|E|$ edges. The adjacency matrix of the graph is denoted by $\bA=(a_{ij})\in \mathbb{R}^{n\times n}$, whose entries $a_{ij}=1$ if $(\nu_i,\nu_j)\in E$, and $a_{ij}=0$ otherwise.
Each node's own feature vector is denoted by $\bx_i\in \mathbb{R}^d$, $i\in[n]$, where $d$ is the dimension of the node feature. Let $\XX\in \mathbb{R}^{n\times d}$ denote the node feature matrix with each row being $d$ features.
The $1$-hop neighborhood of a node $\nu_i$ is defined as the set $\{\nu_j, (\nu_i,\nu_j)\in E\}$, and denote by $N_{(i)}$ the set that includes the node $\nu_i$ and all nodes belonging to its $1$-hop neighborhood.
The main task of graph models is to combine the feature information and the edge information to perform learning tasks.

For an undirected graph, its Laplacian matrix $\bL\in \mathbb{R}^{n\times n}$ is defined as $\bL:=\bD-\bA$, where $\bD \in \mathbb{R}^{n\times n}$ is a degree diagonal matrix whose diagonal entry $d_{ii}=\sum_{j}a_{ij}$ for $i\in [n]$. The semi-definite matrix $\bL$ has an eigen-decomposition written by $\bL=\bU\bLambda\bU^T$, where the columns of $\bU$ are the eigenvectors of $\bL$ and the diagonal entries of diagonal matrix $\bLambda$ are the non-negative eigenvalues of $\bL$.

For a fixed function $g$, we define a graph filter $g(\bL)\in \mathbb{R}^{n\times n}$ as a function on the graph Laplacian $\bL$. Following the eigen-decomposition of $\bL$, we get $g(\bL)=\bU g(\bLambda)\bU^T$, where the eigenvalues are given by $\lambda_i^{(g)}=\{g(\lambda_i),1\leq i\leq n\}$.
We define $\lambda^{max}_G=\max\{|\lambda_i^{(g)}|\}$ as the largest absolute eigenvalue of the graph filter $g(\bL)$.

In this paper, we are concerned with node-level semi-supervised learning problems over the graph $G$. Let $ \mathcal{X}\subset\RR^d$ be the input space in which the node feature is well defined, and accordingly $\mathcal{Y}\subset\mathbb{R}$ be the output space. In the semi-supervised setting, one assumes that only a portion of the training samples are labeled while amounts of unlabeled data are collected easily. Precisely, we merely collect a training set with labels $D=\{\bz_i=(\bx_i,y_i)\}_{i=1}^m$ with $m\ll n$. For statistical inference, one often assumes that these pairwise sample are independently drawn from a joint distribution $\rho$ defined on $\mathcal{X}\times \mathcal{Y}$. In such case, our studied model belongs to node-focused tasks on graph, as opposed to graph-focused tasks where the whole graph can be viewed as a single sample.

The most simple graph neural network, known as the Vanilla GCN, was proposed in \cite{kipf2016semi}, where each layer of a multilayer network is multiplied by the graph filter before applying a nonlinear activation function. 
In a matrix form, a conventional multi-layer GCN is represented by a layer-wise propagation rule
\begin{eqnarray}
    \bH^{(k+1)}=\sigma\big(g(\bL)\bH^{(k)}\bW^{(k+1)}\big),
\end{eqnarray}
where $\bH^{(k+1)}\in \mathbb{R}^{n\times m_{k+1}}$ is the node feature representation output by the $(k+1)$-th GCN layer, and specially $\bH^{(0)}=\XX$ and $m_0=d$. $\bW^{(k+1)}\in \mathbb{R}^{m_k\times m_{k+1}}$ represents the estimated weight matrix of the $(k+1)$-th GCN layer, and $\sigma$ is a point-wise nonlinear activation function. Under the context of GCN for performing semi-supervised learning, the sampling procedure of nodes from the graph $G$ is conducted by two stages. We assume node data are sampled in an i.i.d. manner by first choosing a sample $\bx_i$ or $\bz_i$ at node $i$, and then extracting its neighbors from $G$ to form an ego-graph.

To interpret learning mechanism clearly for GCN, this paper focuses on a node-level task over graph with a single layer GCN model. In such case, putting all graph nodes together, the output function can be written in a matrix form as follows,
\begin{equation}\label{singlegraph}
f(\XX,\bw)=\sigma\big(g(\bL)\XX\bw\big),
\end{equation}
where $\bw\in \mathbb{R}^d$ and $f(\XX,\bw) \in \mathbb{R}^n$.
Some commonly used graph filters include a linear function of $\bA$ as $g(\bA)=\bA+\bI$ \cite{xu2018powerful} or a Chebyshev polynomial of $\bL$ \cite{defferrard2016convolutional}.

Under the context of ego-graph, each node contains the complete information needed for computing the output of a single layer GCN model. Given node with the feature $\bx$, let $N_\bx$ denote a set of the neighbor indexes at most $1$-hop distance neighbors, which is completely determined by the graph filter $g(\bL)$.
Thus we can rewrite the predictor \eqref{singlegraph} for a single node prediction as
\begin{equation}\label{hypothespace}
    f(\bx,\bw)=\sigma\big({\sum}_{j\in N_\bx}e_{\cdot j}\bx_j^T\bw\big),
\end{equation}
where $e_{\cdot j}=[g(\bL)]_{\cdot j}\in \mathbb{R}$ is regraded as a weighted edge between node $\bx$ and its neighbor $\bx_j$, and particularly it still holds $j \in N_\bx$ if and only if $e_{\cdot j} \neq 0$.

%Let $(X,Y)\in \mathcal{X}\times \mathcal{Y}$ be random variables drawn from an unknown distribution $\mathbb{P}$ 

Let $\ell(\cdot,\cdot)$ be a convex loss function, measuring the difference between a predictor and the true label, a variety of supervised learning problems in machine learning can be formulated as a minimization of the expectation risk,
\begin{equation}\label{gmodel}
    \min_{f\in \mathcal{F}}\mathbb{E}\big[\ell\big(Y,f(X)\big)\big],
\end{equation} 
where $\mathcal{F}$ is a hypothesis space under which an optimal learning rule
is generated. 
The standard regression problems correspond to the square loss given by $\ell(u,v)=(u-v)^2$, and the logistic loss is widely used for classification.
The optimal decision function denoted by $f_0$ is any minimizer of \eqref{gmodel} when $\mathcal{F}$ is taken to be the space of all measurable functions. However, $f_0$ can not be computed directly, due to the fact that $\rho$ is often unknown and $\mathcal{F}$ is too complex to compute it possibly. Instead, 
a frequently used method consists of minimizing a regularized empirical risk over a computationally-feasible space
\begin{eqnarray}
    \min_{f\in \mathcal{H}}\frac{1}{m}{\sum}_{i=1}^m\ell\big(y_i,f(\bx_i)\big)+\Omega(f),
\end{eqnarray}
where $\Omega(f)$ is a penalty function that regularizes the complexity of the function $f$, while
%all the observations $\{(y_i,\bx_i)\}_{i=1}^m$ are drawn independently from $\mathbb{P}$, and 
$\mathcal{H}$ is the space of all predictors which needs to be parameterized explicitly or implicitly. 
For instance, the neural network is known as an efficient parameterized approximation to any complex nonlinear function.
In the work, all the functions within $\mathcal{H}$ consist of the form in \eqref{hypothespace}.

\subsection{Empirical Risk Minimization with $\ell_p$-Regularizer}
A fundamental class of learning algorithms can be described as the regularized empirical risk minimization problems. This paper considers an $\ell_p$-regularized learning approach for training the parameters of GCN:
\begin{align}\label{erm}
    \widehat \bw\in \arg \min_{\bw}\frac{1}{n}{\sum}_{i=1}^n\ell\big(y_i,f(\bx_i,\bw)\big)+\lambda\|\bw\|_{\ell_p}^p,
\end{align} 
where $1<p\leq 2$ and the $\ell_p$-norm on $\mathbb{R}^d$ is defined as $\|\bw\|_{\ell_p}^p=\sum_{j=1}^d|w_j|^p$.
For any $1<p'\leq p\leq 2$, there always holds $\|\bw\|_{\ell_p}\leq \|\bw\|_{\ell_{p'}}$, which means that any learning method with the $\ell_{p'}$-regularizer imposes heavier penalties for the parameters than the one with the $\ell_{p}$-regularizer. Specially when $p\rightarrow 1$, the corresponding $\ell_p$-regularized algorithm tends to generate so called soft sparse solutions \cite{koltchinskii2009sparsity}. In contrast to this, the commonly-used $\ell_2$-regularization tends to generate smooth but non-sparse solutions.

Note that we do not require that the minimizer of \eqref{erm} is unique, catering to non-convex problems.
The following lemma tells us that any global minimizer of \eqref{erm} can be upper bounded by a quantity that is inversely proportional to $\lambda$. This simple conclusion is useful in subsequent sections for designing a constrained SGD and our theoretical analysis with respect to the function $\|\bw\|_{\ell_p}^p$.
\begin{lem}\label{bound}
Assume that $\ell(y,\sigma(0))\leq B$ with some $B>0$.	For any $\lambda>0$, any global minimizer $\widehat \bw$ of \eqref{erm} satisfies $\|\widehat \bw\|_{\ell_p}^p\leq {B}/{\lambda}$, and furthermore $\|\widehat \bw\|_{\ell_2} \leq \big({B}/{\lambda}\big)^{1/p}$.
\end{lem}
    \begin{proof}[Proof of Lemma \ref{bound}.]
Since $\widehat \bw$ is a global minimizer of the objective function in \eqref{erm}, this follows that
\begin{align*}
    \lambda\|\widehat \bw\|_{\ell_p}^p&\leq \frac{1}{n}{\sum}_{i=1}^n\ell\big(y_i,f(\bx_i,\bzero)\big)+\lambda\|\bzero\|_{\ell_p}^p
    \leq B.
\end{align*}
Moreover, for any $1<p\leq 2$, it is known that $\|\bw\|_{\ell_2}\leq \|\bw\|_{\ell_p}$ for any $\bw\in \mathbb{R}^d$.
This completes the proof of the lemma.
\end{proof}

Lemma \ref{bound} implies that the empirical solutions of \eqref{erm} are in an $\ell_p$-ball of certain radius which depends on the regularized parameter $\lambda$. This shows that it suffices to analyze statistical behaviors of the given estimator projected into this ball.

%Consider a function $h:\mathbb{R}\rightarrow \mathbb{R}$ defined as $h(\theta)=|\theta|^p$ $(1<p\leq 2)$. Note that we have
%$$
%h'(\theta)=p.\hbox{sign}(\theta).|\theta|^{p-1}.
%$$ 
%Obviously the first derivative of $g$ exists and is continuous for all $\theta\in \mathbb{R}$. Then the gradient of
%$\|\bw\|_{\ell_p}^p$ is written as 
%$$\nabla \|\bw\|_{\ell_p}^p=p\big(\hbox{sign}(w_1).|w_1|^{p-1},\hbox{sign}(w_2).|w_2|^{p-1},...,
%\hbox{sign}(w_d).|w_d|^{p-1}\big).
%$$
\section{Regularized Stochastic Algorithm}
In order to effectively solve non-convex problems with massive data, practical algorithms for machine learning are increasingly constrained to spend less time and use less memory, and can also escape from saddle points that often appear in non-convex problems and tend to converge to a good stationary point. Stochastic gradient descent (SGD) is perhaps the simplest and most well studied algorithm that enjoys these advantages. 
The merits of SGD for large scale learning and the associated computation versus statistics tradeoffs is discussed in detail by the seminal work of \cite{bottou2007tradeoffs}.

A standard assumption to analyze SGD in the literature is that the derivative of the objective function is Lipschitz smooth \cite{verma2019stability,hardt2016train}, however, the $\ell_p$-regularized learning does not meet such condition. To address this issue, we propose a new SGD for \eqref{erm} with an inexact proximal operator, and then develop a novel theoretical analysis for an upper bound of uniform stability \cite{bousquet2002stability}, which is an algorithm-dependent sensitivity-based measurement used for characterizing generalization performance in learning theory.

Given that a positive pair $(p,q)$ satisfies the equality $\nicefrac{1}{p}+\nicefrac{1}{q}=1$, then the norms $\|\bw\|_p$ and $\|\bw\|_q$ are dual to each other. Moreover, the pair of functions $(1/2)\|\bw\|^2_p$ and $(1/2)\|\bw\|^2_q$ are conjugate functions of each other. As a consequence, their gradient mappings are a pair of inverse mapping. Formally, let $p\in(1,2]$
and $q=p/(p-1)$, and define the mapping $\Phi: \mathrm{E}\rightarrow \mathrm{E}^*$ with 
$$\Phi_j(\bw)=\nabla_j\big(\frac{1}{2}\|\bw\|^2_p\big)=\frac{\hbox{sgn}(w_j)|w_j|^{p-1}}{\|\bw\|_p^{p-2}},\,j=1,2,...,d,$$ 
and the inverse mapping $\Phi^{-1}: \mathrm{E}^*\rightarrow \mathrm{E}$ with 
 $$\Phi^{-1}_j(\bv)=\nabla_j\big(\frac{1}{2}\|\bv\|^2_q\big)=\frac{\hbox{sgn}(v_j)|v_j|^{q-1}}{\|\bv\|_q^{q-2}},\,j=1,2,...,d.$$ 
The above conjugate property on $\ell_p$-space and $\ell_q$-space is very useful for bounding uniform stability without the help of strong smoothness, while the latter is a standard assumption in optimization. 
 
\subsection{SGD with Inexact Proximal Operator}

We write $L_i(\bw):=\ell(y_i,f(\bx_i,\bw))$ for notational simplicity, and define a local quadratic approximation of $L_i$ at point $\bw_{D,t}$ as
\begin{align*}
P_{i,r_{i}}(\bw,\bw_{D,t}):=&L_i(\bw_{D,t})+\langle \bw-\bw_{D,t},\nabla L_i(\bw_{D,t})\rangle_2 \\
&+r_{i}\|\bw-\bw_{D,t}\|_2^2.
\end{align*}
At each iteration $t$, let $i_t$ be a random index sampled uniformly from $[n]$ on $D$. Then
replacing $\frac{1}{n}\sum_{i=1}^nL_i(\bw)$ in \eqref{erm} by the quadratic term $P_{i_t,r_{i_t}}(\bw,\bw_{D,t})$, we propose an inexact proximal method for SGD with the $\ell_p$-regularizer. Precisely, we are concerned with
the following iterative to update $\bw_{D,t}$ for a regularized-based SGD,
\begin{align}\label{sgd}
 \min_{\bw}P_{i_t,r_{i_t}}(\bw,\bw_{D,t})+\lambda_t \|\bw\|_{\ell_p}^p.
\end{align}
 In view of the boundedness of $\|\widehat \bw\|_{\ell_2}$ given in Lemma \ref{bound}, we adopt the projection technique to execute a constrained SGD over the empirical risk term in \eqref{sgd}. To this end, we define the projection onto a set $\mathcal{C}$ by
$$
\Pi_{\mathcal{C}}(\bv):=\arg \min_{\bw\in \mathcal{C}}\|\bw-\bv\|_2.
$$ 
This reveals that the definition of projection is an optimization problem in itself. In our case, the set we adopt is given as 
$$\mathcal{C}:=\mathcal{C}_\lambda=\big\{\bw \in \mathbb{R}^d,\,\,\|\bw\|_2\leq (B/\lambda)^{1/p}\big\}.
$$ 
It is well known that, if $\mathcal{C}=\mathcal{B}_2(1)$,
i.e., the unit $\ell_2$ ball, then projection is equivalent to a normalization step
\begin{equation*}
\Pi_{\mathcal{B}_2(1)}(\bv)=
\begin{cases}
\bv/\|\bv\|_2& \text{if} \,\,\|\bv\|_2>1,\\
\bv& \text{otherwise}.
\end{cases}
\end{equation*}

Up to the terms that do not depend on $\bw$, summing
the objective function in \eqref{sgd} and the projection formulate our proposed algorithm as follows
\begin{align}
    \bv_{D,i_t}&=\Pi_{\mathcal{C}_\lambda}\big(\bw_{D,t}-(\eta\nabla L_{i_t}(\bw_{D,t})\big),\label{project}\\
    \bw_{D,t+1}&=\arg \min_{\bw}\big\{\frac{1}{2}\big\|\bw-\bv_{D,i_t}\big\|_2^2+\lambda_t \|\bw\|_{\ell_p}^p\big\},
\label{inexact}
\end{align}
 where $\eta>0$ is the learning rate that depends on $r_{i_t}$, and the $\lambda_t$ may depend on 
 $\lambda$ and $r_{i_t}$.
 
\begin{remark}
The update rule in (\ref{inexact}) is seen as a contraction of conventional SGD, see Lemma \ref{contrative} below for details.
We will obtain analytical solutions of (\ref{inexact}) for some specific $p$ (e.g. $p=1,2$).
Although there is no analytical solutions for general $1<p\leq 2$, the objective function in (\ref{inexact}) is strongly convex over bounded domains and thus a global convergence can be guaranteed.
For the ease of notation, we still denote by $\bw_{D,t+1}$ the realized numerical solution with ignoring the inner optimization error.
\end{remark}

\begin{lem}\label{contrative}
For $1<p\leq 2$ and a vector $\bv \in \mathbb{R}^d$ is given, we define
\begin{align} \label{proxsolu}
\bw^*=	\arg \min_{\bw}\big\{\frac{1}{2}\|\bw-\bv\|_2^2+\lambda \|\bw\|_{\ell_p}^p\big\}:=\hbox{Pro}_{\lambda,p}(\bv).
\end{align}
Then, we conclude that 
$$
|w_j^*|\leq \min\{|v_j|, (|v_j|/(\lambda p))^{1/(p-1)}\}, \quad \,\forall \, j=1,2...,d.
$$
\end{lem}

We defer the proof of Lemma \ref{contrative} to the Appendix. 
Applying Lemma \ref{contrative} and the projection onto $\mathcal{C}_\lambda$ in \eqref{project}, we have the following inequality
\begin{align}\label{boundpara}
\|\bw_{D,t}\|_2\leq (B/\lambda)^{1/p},\quad \forall \,t,\lambda>0.
\end{align}

Consider a function $h:\mathbb{R}\rightarrow \mathbb{R}$ defined as $h(\theta)=|\theta|^p$ $(1<p\leq 2)$. Note that it holds
$$
h'(\theta)=p.\hbox{sign}(\theta).|\theta|^{p-1}.
$$ 
Obviously the first derivative of $h$ exists and is continuous for all $\theta\in \mathbb{R}$. However, we notice that the inexact proximal operator \textit{$\hbox{Pro}_{\lambda,p}$} is not Lipschitz,
due to the fact that $\|\cdot\|_{\ell_p}^p$ is not strongly smooth. Hence, as mentioned earlier, those conventional technique analysis under strongly smooth condition for objective functions are no longer valid in our case.
Fortunately, the function $h$ is strongly convex over bounded domains, as shown in Lemma \ref{bound}, which enables us to avoid restrictive smooth assumptions by an alternative proof strategy.

\begin{figure*}[ht]
	\centering
	\includegraphics[width=0.95\linewidth]{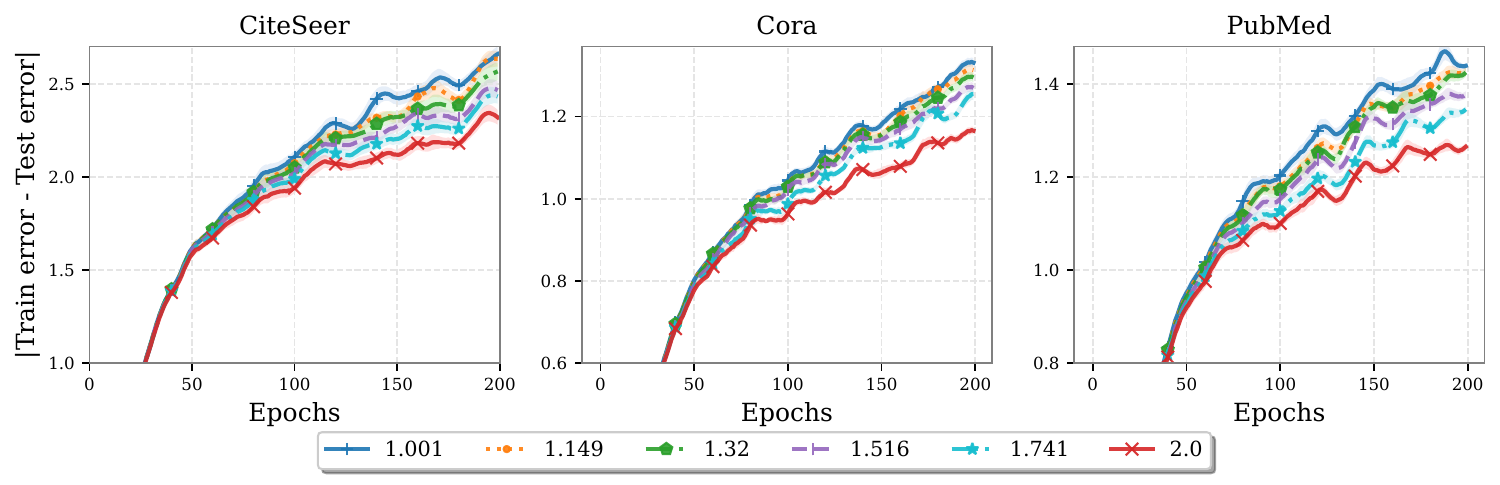}
	% \vspace{-0.5em}
	\caption{\textbf{Generalization Gaps as a function of the number of epochs under various $p$ on three datasets.} We observe that the miniature $p$ achieves weaker generalization gap and thus worse than the significant $p$.}
	\label{fig:generalization_gap}
\end{figure*}

\section{Stability and Generalization Bounds}
In this section, we provide algorithm-dependent generalization bounds via the notion of stability for $\ell_p$-regularized GCN.
To this end, we first introduce the notion of algorithmic stability and thereby present a generalization bound associated with the algorithmic stability. 

Let $\mathcal{A}_D$ be a learning algorithm trained on dataset $D$, which can be viewed as a map from 
$D\rightarrow \mathcal{H}$. For GCN, we set $\mathcal{A}_D=f(\bx,\bw_D)$. The overall learning performance of
$\mathcal{A}_D$ is measured by the following expected risk:
%The expected risk of
%$\mathcal{A}_D$ is defined as 
$$
R(\mathcal{A}_D):=\mathbb{E}_{\bz}\big[\ell\big(y,f(\bx,\bw_D)\big)\big].
$$
%Since the joint distribution $\rho$ on $\bz$ is often unknown, we consider 
Accordingly, the empirical risk of $\mathcal{A}_D$ with the loss $\ell$ is given as 
$$
R_n(\mathcal{A}_D):=\frac{1}{n}{\sum}_{i=1}^n\ell\big(y_i,f(\bx_i,\bw_D)\big).
$$
Even when the sample is fixed, $\mathcal{A}_D$ may be still a randomized algorithm due to the randomness of algorithm procedure (e.g. SGD). In this context, we define the expected generalization error as 
$$
E_{\text{gen}}:=\mathbb{E}_{\mathcal{A}}\big[R(\mathcal{A}_D)-R_n(\mathcal{A}_D)\big],
$$
where the expectation $\mathbb{E}_{\mathcal{A}}$ is taken over the inherent randomness of $\mathcal{A}_D$. 

For a randomized algorithm, to introduce the notation of its uniform stability, we need to define two datasets as follows. Given the training set $D$ defined as above, we introduce two related sets in the following:

Removing the $i$-th data point in the set $D$ is represented as
$$
D^{\backslash i}=\{\bz_1,\bz_2,...\bz_{i-1},\bz_{i+1},...,\bz_n\},
$$
and replacing the $i$-th data point in $D$ by $\bz^{'}_i$ is represented as
$$
D^{i}=\{\bz_1,\bz_2,...\bz_{i-1},\bz^{'}_i,\bz_{i+1}...,\bz_n\}.
$$
\begin{defn}
	A randomized learning algorithm $\mathcal{A}_D$ is $\beta_n$-uniformly stable with respect to a loss function
	$\ell$, if it satisfies
	$$
	\sup_{D,\bz}\big|\mathbb{E}_{\mathcal{A}}[\ell(y,f(\bx,\bw_D))]-
	\mathbb{E}_{\mathcal{A}}[\ell(y,f(\bx,\bw_{D^{\backslash i}}))]\big|\leq \beta_n.
	$$
\end{defn}
By the triangle inequality, the following result on another uniform stability associated with $S^{i}$ holds
$$
\sup_{D,\bz}\big|\mathbb{E}_{\mathcal{A}}[\ell(y,f(\bx,\bw_D))]-
\mathbb{E}_{\mathcal{A}}[\ell(y,f(\bx,\bw_{D^{ i}}))]\big|\leq 2\beta_n.
$$
Stability is property of a learning algorithm, roughly speaking, if two training samples are close to each other, a stable algorithm will generate close output results. There are many variants of stability, such as hypothesis stability \cite{kearns1999algorithmic}, sample average stability \cite{shalev2010learnability} and 
uniform stability. This paper will focus on the uniform stability, since it is closely related to other types of stability. 

The following lemma shows that a randomized learning algorithm with uniform stability can guarantee meaningful
generalization bound, which has been proved in \cite{verma2019stability}.
\begin{lem}\label{generviasta}
	A uniform stable randomized algorithm $(\mathcal{A}_D,\beta_n)$ with a bounded loss function
	$0\leq \ell(y,f(\bx))\leq B$, satisfies the following generalization bound with probability at least
	$1-\delta$, over the random draw of $D,\bz$ with $\delta\in(0,1)$,
	$$
	\mathbb{E}_{\mathcal{A}}\big[R(\mathcal{A}_D)-R_n(\mathcal{A}_D)\big]\leq 2\beta_n+(4n\beta_n+B)\sqrt{\frac{\log(1/\delta)}{2n}}.
	$$
\end{lem}

We now give some smooth assumptions on loss function and activation function used for analyzing the stability of stochastic gradient descent.
The following assumptions are very standard in optimization literature.
\begin{assumption}[Smoothness for loss function and activation function] \label{assum:smooactiva}
	We assume that the loss function is lipschitz continuous and smooth,
	\begin{align*}
		& |\ell(y,f(\cdot))-\ell(y,f'(\cdot))|\leq a_{\ell}|f(\cdot)-f'(\cdot)|.\quad \forall\, f,f'\in \mathcal{H}\\
		& |\ell'(y,f(\cdot))-\ell'(y,f'(\cdot))|\leq b_{\ell}| f(\cdot)- f'(\cdot)|,
	\end{align*}	
	where $a_\ell$ and $b_\ell$ are two positive constants.
	Similarly, the activation function also satisfies
	\begin{align*}	
		|\sigma(x)-\sigma(y)|\leq a_{\sigma}|x-y|, \,\,
		&|\sigma'(x)-\sigma'(y)|\leq b_{\sigma}|x-y|\, \\
		\text{and}\,\, &\ell(y,f(\bx))\leq B,\,\forall\,x,y\in \mathbb{R},
	\end{align*}
	where $a_\sigma,\,b_\sigma$ and $B$ are positive constants as well.
\end{assumption}
We now present an explicit stability bound for GCN with $\ell_p$-regularizer via stochastic gradient method. 

\begin{thm}\label{stablitybeta}
	Suppose that the loss and activation functions are Lipschitz-continuous and smooth functions (Assumption \ref{assum:smooactiva}). Then a single layer GCN model, trained by the proposed SGD given in \eqref{project}-\eqref{inexact} for iteration $T$, is $\beta_n$-uniformly stable, precisely
\begin{align*}
\beta_n\leq a_{\ell}^2a_{\sigma}^2\lambda^{max}_{G}  \frac{ \eta C_{p,\lambda}\bg_{e}}{n}\sum_{t=1}^T\Big(C_{p,\lambda}\big(1+(a_\sigma^2+a_\ell)\eta\bg_{e}^2\big)\Big)^{t-1},
\end{align*}	
where $C_{p,\lambda}:=\frac{28}{p(p-1)\lambda_t}\big(\nicefrac{B}{\lambda}\big)^{(3-p)/p}$ and $\bg_{e}:=\sup_{\bx}\big\|\sum_{j\in N_\bx}e_{\cdot j}\bx_j\big\|_2$.
\end{thm}

The proof procedure of Theorem \ref{stablitybeta} will be given in Appendix. The key step of our proof is that in such a scenario, the error caused by the difference in the nonconvex empirical risk of GCN grows polynomially with the number of iterations.
\begin{remark}
    Theorem \ref{stablitybeta} provides the uniform stability for the last iterate of SGD for $\ell_p$-regularized GCN. It is worth mentioning that the upper bound of this stability depends on the graph structure (i.e. $\bg_{e}$ and $\lambda_G^{max}$) and the regularized hyper-parameter $p$, as well as the sample size and the learning rate in SGD. 
\end{remark}
\begin{remark}
    More precisely, the result in Theorem \ref{stablitybeta} shows that the stability bound decreases inversely with the scale of $p$. It increases as the graph structured parameter $\lambda_G^{max}$ increases.
\end{remark}

Substituting Lemma \ref{generviasta} into Theorem \ref{stablitybeta} above, we obtain the generalization bounds with uniform stability for a single layer GCN with $\ell_p$-regularizer.

\begin{thm}\label{genera}
Under the same conditions as Theorem \ref{stablitybeta}. with a high probability, the following generalization bound holds 
\begin{align*}
    &\mathbb{E}_{\mathcal{A}}\big[R(\mathcal{A}_D)-R_n(\mathcal{A}_D)\big] \\
    &= \mathcal{O}\Big( \lambda^{max}_{G}  \frac{ \eta C_{p,\lambda}\bg_{e}}{\sqrt{n}}{\sum}_{t=1}^T\big(C_{p,\lambda}\big(1+(a_\sigma^2+a_\ell)\eta\bg_{e}^2\big)\big)^{t-1}\Big).
\end{align*}
\end{thm}

\begin{remark}
Based on the result of Theorem  \ref{genera},  we conclude that $\ell_p$-regularization for $1 < p \leq 2$ generalizes.
Note that, when $p\rightarrow 1$, the stability bound breaks due to the sparsity of $\ell_1$-regularization.
The smaller $p$ becomes, the greater the stability parameter $\beta_n$ is, but at the same time the obtained parameter $\widehat{\bw}$ in \eqref{erm}  tends to be sparse, shown in \cite{koltchinskii2009sparsity}. These
properties are also verified in our experimental evaluation.
\end{remark}

\begin{remark}
Although a small learning rate means a smaller generalization gap, the parameter range in which this SGD searches will be very small, resulting in a larger training error. Such conclusion is also applicable to various SGD for general models.
\end{remark}

\begin{table*}[t]
    \centering
    \scalebox{0.9}{
    \begin{tabular}{l | c | c c c c c c}
        \toprule
        Dataset & Graph Filter & ${p=1.001}$ & ${p=1.149}$   & ${p=1.320}$   &${p=1.516}$  & ${p=1.741}$ &${p=2}$ \\
        
        \midrule
        \multirow{4}{*}{Citeseer}
        &Augmented Normalized	&$\mathbf{57.44\pm0.87}$	&$54.69\pm0.82$ &$52.41\pm0.73$	&$50.72\pm0.67$	&$50.27\pm0.68$	&$50.18\pm0.69$ \\
        &Normalized	    &$\mathbf{57.02\pm1.32}$	&$54.47\pm0.99$	&$51.93\pm0.66$ &$50.39\pm0.62$	&$49.95\pm0.62$	&$49.88\pm0.63$ \\
        &Random Walk	&$\mathbf{56.17\pm1.49}$	&$53.84\pm1.20$	&$51.84\pm0.84$ &$50.54\pm0.78$	&$50.16\pm0.81$	&$50.13\pm0.8$ \\
        &Unnormalized	&$\mathbf{60.62\pm2.25}$	&$58.78\pm2.31$	&$57.52\pm2.23$ &$56.61\pm2.25$	&$56.32\pm2.3$	&$56.45\pm2.2$ \\
        
        \midrule
        \multirow{4}{*}{Cora}
        &Augmented Normalized	&$\mathbf{56.66\pm3.43}$	&$54.46\pm2.68$	&$52.18\pm2.13$ &$50.79\pm1.86$	&$50.41\pm1.77$	&$50.27\pm1.76$ \\
        &Normalized	    &$\mathbf{55.77\pm3.68}$	&$53.74\pm2.78$	&$51.6\pm2.10$	&$50.22\pm1.82$	&$49.77\pm1.82$	&$49.69\pm1.83$ \\
        &Random Walk	&$\mathbf{53.43\pm2.12}$	&$52.03\pm1.94$	&$51.08\pm1.37$	&$50.29\pm1.18$	&$50.08\pm1.18$	&$50.07\pm1.17$ \\
        &Unnormalized	&$\mathbf{64.94\pm3.61}$	&$63.86\pm3.5$	&$63.31\pm3.59$	&$62.79\pm3.53$	&$62.87\pm3.56$	&$62.82\pm3.5$ \\
        
        \midrule
        \multirow{4}{*}{Pubmed}
        &Augmented Normalized	&$\mathbf{54.13\pm2.16}$	&$52.61\pm2.62$	&$50.83\pm1.92$	&$49.73\pm1.67$	&$49.24\pm1.69$	&$49.06\pm1.63$ \\
        &Normalized	    &$\mathbf{53.27\pm3.40}$	&$52.48\pm2.35$	&$51.28\pm2.09$	&$50.29\pm1.96$	&$49.56\pm1.78$	&$49.30\pm1.55$ \\
        &Random Walk	&$\mathbf{54.72\pm3.57}$	&$53.07\pm3.20$	&$51.82\pm2.94$	&$51.03\pm2.84$	&$50.85\pm2.8$ 	&$50.77\pm2.76$ \\
        &Unnormalized	&$\mathbf{74.54\pm6.26}$	&$74.10\pm6.40$	&$73.29\pm6.3$	&$73.25\pm6.57$	&$72.91\pm6.12$	&$73.09\pm5.96$ \\
        \bottomrule
    \end{tabular}
    }
    % \vspace{-0.5em}
    \caption{The sparsity ($\%$) of obtained parameter $\widehat{\bw}$ under different $p$ and normalization steps on three datasets.}
    \label{tab:sparsity}
\end{table*}

\section{Experiments} \label{experiment}
In this section, we conduct experiments to validate our theoretical findings. In particular, we show that there exists a stability-sparsity trade-off with varying $p$, and the uniform stability of our GCN depends on the largest absolute eigenvalue of its graph filter. To do it, we first introduce the experimental settings. Then we assess the uniform stability of GCN with semi-supervised learning tasks under varying $p$. Finally, we evaluate the effect of different graph filters on the GCN stability bounds.

\begin{figure*}[ht]
	\centering
	\includegraphics[width=0.95\linewidth]{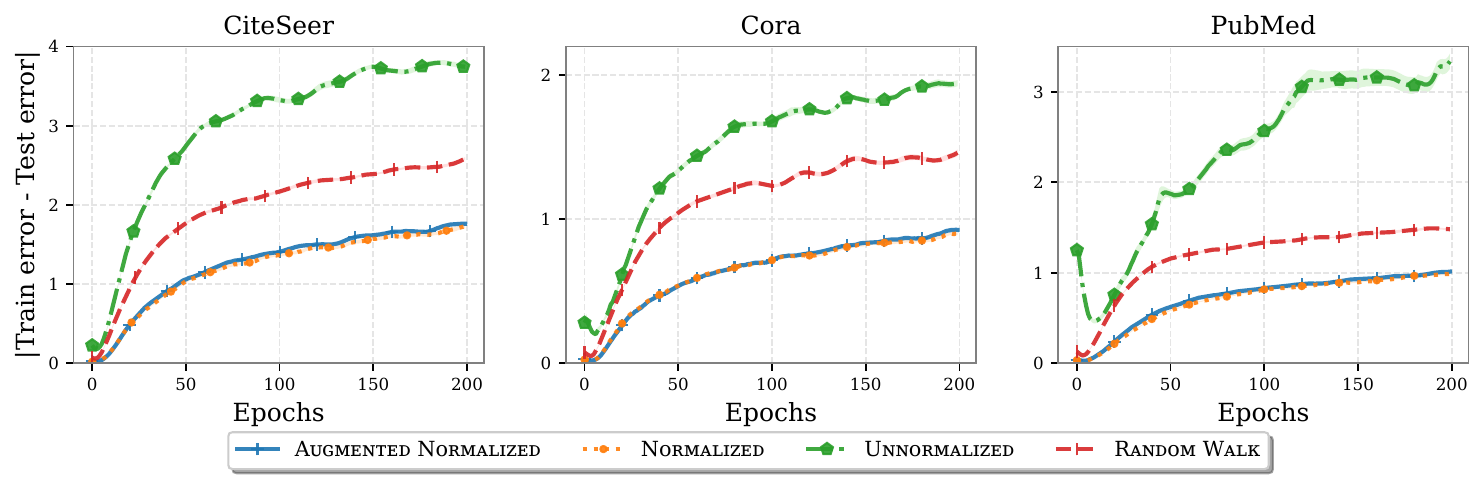}
	% \vspace{-0.5em}
	\caption{\textbf{Generalization Gaps under different normalization steps on three datasets.} We observe that the normalized filters yield significantly lower generalization gaps than the unnormalized and random walk filters. }
	\label{fig:3}
\end{figure*}

\subsection{Experimental Setup}
\paragraph{Datasets.}
We conduct experiments on three citation network datasets: Citeseer, Cora, and Pubmed \cite{sen2008collective}. In every dataset, each document is represented as spare bag-of-words feature vectors. The relationship between documents consists of a list of citation links, which can be treated as undirected edges and help construct the adjacency matrix. These documents can be divided into different classes and have the corresponding class label.

\paragraph{Baselines.}
We implement several empirical experiments with a representative GCN model \cite{kipf2016semi}. For all datasets, we use 2-layer neural networks with 16 hidden units. In all cases, we evaluate the difference between the learned weight parameters and the generalization gap of two GCN models trained on datasets $D$ and $D^i$, respectively. 
Specifically, we generate $D^i$ by choosing a random data point in the training set $D$ and altering it with a different random point. We also record the training and testing errors gap and the parameter distance $\sqrt{\|\widehat{\bw}-\widehat{\bw}^{'}\|^2/(\|\widehat{\bw}\|^2+\|\widehat{\bw}^{'}\|^2)},$ where $\widehat{\bw}$ and $\widehat{\bw}^{'}$ are the weight parameters of the respective models per epoch.

\paragraph{Training settings.}
For each experiment, we initialize the parameters of GCN models with the same random seeds and then train all models for a maximum of 200 epochs using the proposed Inexact Proximal SGD. We repeat the experiments 10 times and report the average performance as well as the standard variance. For all methods, the hyperparameters are tuned from the following search space: (1) learning rate: \{1, 0.5, 0.1, 0.05\}; (2) weight decay: 0; (3) dropout rate: $\{0.3,0.5\}$; (4) regularization parameter $\lambda$ is set to 0.001. 

\subsection{The Effect of Varying $p$}
\paragraph{Generalization gap.}
In this experiment, we empirically compare the effect of $p$ on the GCN stability bounds using different $p \in \{1.001, 1.149, 1.32, 1.516, 1.741, 2\}$. 
We quantitatively measure the generalization gap defined as the absolute difference between the training and testing errors and the difference between learned weights parameters of GCN models trained on $D$ and $D'$, respectively. 
It can be observed that in Figure \ref{fig:generalization_gap}, these empirical observations are in line with our stability bounds (see also results in Figure 3 in Appendix).%and \ref{fig:parameter_diff}

\paragraph{Sparsity.} 
Besides the convergence, we have a particular concern about the sparsity of the solutions. The sparsity ratios for $l_p$-based method are summarized in Table \ref{tab:sparsity}. Observe that $\ell_p$-regularized learning with $p \rightarrow 1$ identifies most of the sparsity pattern but behaves much worse in generalization.

\subsection{The Effect of Graph Filters}
\paragraph{Different normalizations steps.}
In this experiment, we mainly consider the implication of our results in following designing graph convolution filters:
(1) Unnormalized Graph Filters: $g(\bL)=\bA+\mathbf{I}$;
(2) Normalized Graph Filters: $g(\bL)=\mathbf{D}^{-1/2}\mathbf{A}\mathbf{D}^{-1/2}+\mathbf{I}$;
(3) Random Walk Graph Filters: $g(\bL)=\mathbf{D}^{-1}\mathbf{A}+\mathbf{I}$;
(4) Augmented Normalized Graph Filters: $g(\bL)=(\mathbf{D} + \mathbf{I})^{-1/2} ( \mathbf{A} + \mathbf{I} ) (\mathbf{D} + \mathbf{I})^{-1/2}$.

In this experiment, we quantitatively measure the generalization gap and parameter distance per epoch. From Figure \ref{fig:3}, it is clear that the Unnormalized Graph Filters and Random Walk Graph Filters show a significantly higher generalization gap than the normalized ones. The results hold consistently across the three datasets. Hence, these empirical results are also consistent with our generalization error bound. We note that the generalization gap and parameter distance become constant after a certain number of iterations. More results can be found in the supplementary material.

\section{Conclusion}
In this paper, we present an explicit theoretical understanding of stochastic learning for GCN with the $l_p$ regularizer and analyze the stability of our regularized stochastic algorithm. In particular, our derived bounds show that the uniform stability of our GCN depends on the largest absolute eigenvalue of its graph filter, and there exists a generalization-sparsity tradeoff with varying $p$. It is worth noting that previous generalization analysis based on stability notation assumed that objectiveness is a second derivative function, which is no longer applicable to our $l_p$-regularized learning scheme. 
To address this issue, we propose a new proximal SGD algorithm for GCNs with an inexact operator, which exhibits comparable empirical performances.

% \clearpage
\section*{Acknowledgements}
Lv's research is supported by the ``QingLan'' Project of Jiangsu Universities. Ming Li would like to thank the support from the National Natural Science Foundation of China (No. U21A20473, No. 62172370) and Zhejiang Provincial Natural Science Foundation (No. LY22F020004).

\section*{Contribution Statement}
Linsen We and Ming Li contributed equally to this work.

\bibliographystyle{named}
\bibliography{ijcai23}

\clearpage
\onecolumn

\begin{appendices}
\crefalias{section}{appendix}
\crefalias{subsection}{appendix}
\crefalias{subsubsection}{appendix}
The appendices are structured as follows. In Appendix \ref{appendix:ProofSketch}, we provide the proof procedure of Theorem \ref{stablitybeta}. The details of proof for Lemma \ref{contrative} is sketched in In Appendix \ref{appendix:ProofLemmas}. In Appendix \ref{app:Experiments}, we provide additional experiments and detailed setups.

% We provide following details omitted in the main paper:
% \begin{itemize}
%     \setlength{\itemsep}{-0pt}
% 	\setlength{\parsep}{0pt}
% 	\setlength{\parskip}{0pt}
%     \item Appendix \ref{appendix:ProofSketch}: the proof procedure of Theorem \ref{stablitybeta}.
%     \item Appendix \ref{appendix:ProofLemmas}: details of proof for Lemma \ref{contrative}, \ref{convexdiff}, \ref{derivativebound}, \ref{diffboundscase2} and additional lemma.
% \end{itemize}

% \begin{spacing}{1.05}
 \listofappendices
% \end{spacing}
\clearpage

\section{Proof of Theorem}
\subsection{Proof Sketch} \label{appendix:ProofSketch}
\label{proofs}
We describe the main procedures by training a GCN using SGD on datasets $D$ and $D^{i}$ which differ in one data point. Let $Z=\{\bz_1,...,\bz_t,...,\bz_T\}$ be a sequence
of samples, where $\bz_t$ is an i.i.d. sample drawn from
$D$ at the $t$-th iteration of SGD during a training run of the GCN. Training the same GCN using SGD on $D^i$ means
that we endow the same sequence to the GCN except that if $\bz_t=(\bx_i,y_i)$ occurs for some $t$, we replace it
with $\bz_t'=(\bx_i',y'_i)$. We denote this sample sequence by $Z'$. Let $\{\bw_{D,0},\bw_{D,1},...,\bw_{D,T}\}$
and $\{\bw_{D^i,0},\bw_{D^i,1},...,\bw_{D^i,T}\}$ be the corresponding sequences of the weight parameters learning by
running SGD on $D$ and $D^i$, respectively. 

Given two sequences of the weighted parameters, $\{\bw_{D,0},\bw_{D,1},...,\bw_{D,T}\}$
and $\{\bw_{D^i,0}, \bw_{D^i,1},...,\bw_{D^i,T}\}$. At each iteration, we define the difference of these two sequences by 
$$
\Delta\bw_t:=\bw_{D^i,t}-\bw_{D,t}.
$$ 

We now relate $\Delta\bw_t$ to the stability parameter $\beta_n$ given in Definition 1.
Since the loss is Lipschitz continuous by Assumption \ref{assum:smooactiva}, for any $\bz=(\bx,y)$ we have
\begin{align}\label{reduce}
&\big|\mathbb{E}_{\mathcal{A}}[\ell(y,f(\bx,\bw_D))]-
\mathbb{E}_{\mathcal{A}}[\ell(y,f(\bx,\bw_{D^{i}}))]\big|\nonumber\\
\leq & a_{\ell}\mathbb{E}_{\mathcal{A}}[|f(\bx,\bw_D)-f(\bx,\bw_{D^i})|]\nonumber\\
\leq & a_{\ell}a_{\sigma}\mathbb{E}_{\mathcal{A}}\Big[\Big|{\sum}_{j\in N_\bx}e_{\cdot j}\bx_j^T(\bw_D-\bw_{D'})\Big|\Big]\nonumber\\
\leq & a_{\ell}a_{\sigma} \mathbb{E}_{\mathcal{A}}[\|\Delta \bw\|_2]\cdot\big\|{\sum}_{j\in N_\bx}e_{\cdot j}\bx_j\big\|_2 ,
\end{align}
where the second inequality follows from the Lipschitz property of $\sigma(\cdot)$, and the last one is based on the
Cauchy-Schwarz inequality.

As proved in \cite{verma2019stability}, $\bg_{e}=\big\|\sum_{j\in N_\bx}e_{\cdot j}\bx_j\big\|_2$ can be bounded in terms of the largest absolute eigenvalue of the graph convolution filter $g(\bL)$.
Thus, to provide an upper bound of $\beta_n$ in uniform stability, the key point is to bound $\mathbb{E}_{\mathcal{A}}[\|\Delta \bw\|_2]$ involved in \eqref{reduce}.

%{\bf $\ell$-Lipschitz Continuous and Smooth Loss Function}:

\subsection{Bounding $\|\Delta\bw_{t+1}\|_2$}
This subsection is devoted to bounding $\|\Delta\bw_{t+1}\|_2$ due to the randomness inherent in SGD. This will be
proved though a series of five lemmas. Different from existing related work \cite{verma2019stability,hardt2016train}, the strong convexity of $|x|^p$ over bounded domains plays an important role in our desired results. It should be pointed out that, the first derivative of our objective function is not Lipschitz,  the standard technique for proving the stability of SGD \cite{hardt2016train} is no longer applicable to our setting. 

\begin{prop}\label{diffbound}
Suppose that the loss and activation functions are Lipschitz-continuous and smooth functions (Assumption \ref{assum:smooactiva}). Then a single layer GCN model, trained by the proposed algorithm \eqref{project}-\eqref{inexact} at step $t$,  satisfies
\begin{align}\label{iterequ}
\|\Delta\bw_{t+1}\|_2 \leq C_{p,\lambda} \big(\|\Delta\bw_{t}\|_2  
+\eta\|\nabla \ell(y_t,f(\bx_t,\bw_{D,t}))-\nabla \ell(y_t',f(\bx_t',\bw_{D^i,t}))\|_2\big).
\end{align}
At step $t$, $\bz_t=(\bx_t,y_t)$ is a drawn sample from $D$ and $\bz_t'=(\bx_t',y_t')$ is a drawn sample from $D^i$.
Here $C_{p,\lambda}$ is defined as Theorem \ref{stablitybeta}.
\end{prop}

\begin{proof}[Proof of Proposition \ref{diffbound}]
For every $a,b\in \mathbb{R}$, we know the following equality from (4) in \cite{wibisono2009sufficient} that 
\begin{align}\label{strongconvex}
|a|^p+|b|^p-2\big|\frac{a+b}{2}\big|^p=\frac{1}{4}(b-a)^2p(p-1)|c|^{p-2},
\end{align}
where $\min\{a,b\}\leq c\leq \max\{a,b\}$ and   the choice of $c$ in our case will be given later. The equality \eqref{strongconvex} shows that $|x|^p$ $(1<p\leq 2)$ is strongly convex over bounded domains, even though its first derivative  is not Lipschitz smooth.

In particular, we can set $a$ and $b$ to be the $j$-th components $\bw^j_{D,t+1}$ of $\bw_{D,t+1}$ and $\bw^j_{D^i,t+1}$ of $\bw_{D^i,t+1}$ respectively. Then we obtain
\begin{align}\label{trsconvex}
|\bw^j_{D,t+1}|^p+|\bw^j_{D^i,t+1}|^p-2\Big|\frac{\bw^j_{D,t+1}+\bw^j_{D^i,t+1}}{2}\Big|^p
=\frac{1}{4}(\bw^j_{D,t+1}-\bw^j_{D^i,t+1})^2p(p-1)|c_j|^{p-2},
\end{align}
where $\min\{\bw^j_{D,t+1},\bw^j_{D^i,t+1}\}\leq c_j\leq \max\{\bw^j_{D,t+1},\bw^j_{D^i,t+1}\}$. 
As stated in
\eqref{boundpara}, we can derive a bound $c_j$ as follows
$$
|c_j|\leq \max\{|\bw^j_{D,t+1}|,|\bw^j_{D^i,t+1}|\}\leq \Big(\frac{B}{\lambda}\Big)^{1/p}.
$$

Furthermore, since $1<p\leq 2$, this follows that
$$
|c_j|^{p-2}\geq \Big(\frac{B}{\lambda}\Big)^{(p-2)/p},\quad \forall \,j\in [d].
$$

Substituting the bound on $c_j$ to  \eqref{trsconvex} and then summing over $j\in[d]$, we obtain
\begin{align}\label{sumconvex}
&\|\bw_{D,t+1}\|_{\ell_p}^p+\|\bw_{D^i,t+1}\|_{\ell_p}^p-
2\Big\|\frac{\bw_{D,t+1}+\bw_{D^i,t+1}}{2}\Big\|^p_{\ell_p}\nonumber\\
&\geq \frac{1}{4}p(p-1)\Big(\frac{B}{\lambda}\Big)^{(p-2)/p}\|\bw_{D,t+1}-\bw_{D^i,t+1}\|_{\ell_2}^2.
\end{align}

Seen from \eqref{sumconvex}, it suffices to bound the term 
$$
\|\bw_{D,t+1}\|_{\ell_p}^p+\|\bw_{D^i,t+1}\|_{\ell_p}^p-
2\Big\|\frac{\bw_{D,t+1}+\bw_{D^i,t+1}}{2}\Big\|^p_{\ell_p},
$$
which is completed by the following lemma. 
\end{proof}

\begin{lem}\label{convexdiff}
 For any $\theta\in[0,1]$, we have
	\begin{align*}
	&\lambda_t\big(\|\bw_{D,t+1}\|_{\ell_p}^p-\|\bw_{D,t+1}+\theta\Delta\bw_{t+1}\|_{\ell_p}^p +\|\bw_{D^i,t+1}\|_{\ell_p}^p-\|\bw_{D^i,t+1}-\theta\Delta\bw_{t+1}\|_{\ell_p}^p\big)\\
	&\leq 7(B/\lambda)^{1/p}\big(\|\Delta\bw_{t}\|_2+\eta\|\nabla \ell(y_t,f(\bx_t,\bw_{D,t}))
	-\nabla \ell(y_t',f(\bx_t',\bw_{D^i,t}))\|_2 \big).
	\end{align*}
\end{lem}
\begin{proof}
    The proof for Lemma \ref{convexdiff} is provided in Appendix \ref{sub:convexdiff}.
\end{proof}

Now we  proceed with the proof of Proposition \ref{diffbound}.  Applying  Lemma \ref{convexdiff} with  $\theta=\frac{1}{2}$ yields that 
\begin{align*}
&\lambda_t\Big(\|\bw_{D,t+1}\|_{\ell_p}^p+\|\bw_{D^i,t+1}\|_{\ell_p}^p -2\Big\|\frac{\bw_{D,t+1}+\bw_{D^i,t+1}}{2}\Big\|^p_{\ell_p}\Big)\\
& \leq 7\big(\frac{B}{\lambda}\big)^{1/p}\big(\|\Delta\bw_{t}\|_2+\eta\|\nabla \ell(y_t,f(\bx_t,\bw_{D,t})) - \nabla \ell(y_t',f(\bx_t',\bw_{D^i,t}))\|_2 \big).
\end{align*}

Combining the inequality above with \eqref{sumconvex}, we get
\begin{align*}
& \frac{\lambda_t}{28}p(p-1)\big(\frac{B}{\lambda}\big)^{(p-3)/p}\|\bw_{D,t+1}-\bw_{D^i,t+1}\|_{\ell_2}^2\nonumber \leq  \|\Delta\bw_{t}\|_2 +\eta\big\|\nabla \ell(y_t,f(\bx_t,\bw_{D,t}))-\nabla \ell(y_t',f(\bx_t',\bw_{D^i,t}))\big\|_2,
\end{align*}
which immediately yields our desired result in Proposition \ref{diffbound}. 

Since $\bz_t$ and $\bz_t'$ are two random samples from different datasets ($D$ and $D^i$) respectively, we need to consider 
one of the following two scenarios, which must occur at iteration $t$.

\textit{(Scenario 1)} At step $t$, SGD picks a sample 
$\bz_t=\bz_t'=(\bx,y)$ which is identical in $Z$ and $Z'$, occurring with probability $(n-1)/n$.
% In this scenario,
%we shall claims that
%\begin{align}\label{case1bound}
%\|\Delta\bw_{t+1}\|_2\leq C_{p,\lambda} \big(\|\Delta\bw_{t}\|_2+\eta_t\|\nabla \ell(y,f(\bx,\bw_{D,t}))-\nabla \ell(y,f(\bx,\bw_{D^i,t}))\|_2\big),
%\end{align}

\textit{(Scenario 2)} At step $t$, SGD picks the only samples that $Z$ and $Z'$ differs, e.g. $\bz_t\neq \bz_t'$, which occurs with probability $1/n$. 

Consider the {\it Scenario $1$} in Lemma \ref{derivativebound} below. In this case,
we will show  that  $\|\nabla \ell(y,f(\bx,\bw_{D,t}))-\nabla \ell(y,f(\bx,\bw_{D^i,t}))\|_2$ can be bounded by 
the terms $\|\Delta \bw_t\|_2$ and $\bg_{e}^2$. 

\begin{lem}\label{derivativebound}
Under Assumption \ref{assum:smooactiva} for Scenario 1. 
%Suppose that both  $\ell(\cdot)$ and $\sigma(\cdot)$ are the Lipschitz continuous and smooth,
 There holds
\begin{align*}
\big\|\nabla \ell(y,f(\bx,\bw_{D,t}))-\nabla \ell(y,f(\bx,\bw_{D^i,t}))\big\|_2 
\leq (a_\sigma^2+a_\ell) \big\|\sum_{j\in N_\bx}e_{\cdot j}\bx_j\big\|_2^2 \|\Delta \bw_t\|_2.
\end{align*}
\end{lem}
\begin{proof}
The proof for Lemma \ref{derivativebound} is provided in Appendix \ref{sub:derivativebound}.
\end{proof}

Unlike Scenario 1 given in Lemma \ref{derivativebound}, the following lemma gives a more rough bound of the gradient difference under  Scenario 2.

\begin{lem}\label{diffboundscase2}
	Under Assumption \ref{assum:smooactiva} for Scenario 2. 
	%Suppose that both  $\ell(\cdot)$ and $\sigma(\cdot)$ are the Lipschitz continuous and smooth,
	There holds
	\begin{align*}
	\big\|\nabla \ell(y,f(\bx,\bw_{D,t}))-\nabla \ell(y',f(\bx',\bw_{D^i,t}))\big\|_2 \leq 2a_\ell a_\sigma \sup_{\bx}\big\|\sum_{j\in N_\bx}e_{\cdot j}\bx_j\big\|_2.
	\end{align*}
\end{lem}
\begin{proof}
The proof for Lemma \ref{diffboundscase2} is also provided in Appendix \ref{sub:diffboundscase2}. 
\end{proof}

Substituting the results derived in  Lemma \ref{derivativebound} and Lemma \ref{diffboundscase2} to Proposition 
\ref{diffbound}, and taking expectation over all possible sample sequences $\bz,\bz'$ from $D$ and $D^i$, we have the following lemma. 
\begin{lem}\label{recusweight}
Suppose that the loss and activation functions are Lipschitz-continuous and smooth functions (Assumption \ref{assum:smooactiva}). Then a single layer GCN model, trained by the proposed algorithm \eqref{project}-\eqref{inexact} at step $t$,  satisfies
\begin{align*}
\mathbb{E}_{SGD}\big[\|\Delta\bw_{T}\|_2\big] \leq \frac{2a_\ell a_\sigma \eta C_{p,\lambda}\bg_{e}}{n}\sum_{t=1}^T
\Big(C_{p,\lambda}\big(1+(a_\sigma^2+a_\ell)\eta\bg_{e}^2\big)\Big)^{t-1}.
\end{align*}
\end{lem}
\begin{proof}[Proof of Lemma \ref{recusweight}.]
From \eqref{iterequ} in Proposition \ref{diffbound}, this together with   the derived results in    Lemma \ref{derivativebound} and Lemma \ref{diffboundscase2} yields that, $\mathbb{E}_{SGD}\big[\|\Delta\bw_{t+1}\|_2\big]$ can be upper bounded by
\begin{align*}
& C_{p,\lambda}\Big(\mathbb{E}_{SGD}\big[\|\Delta\bw_{t}\|_2\big]+\mathbb{E}_{SGD}\big[\eta\|\nabla \ell(y_t,f(\bx_t,\bw_{D,t})) -\nabla \ell(y_t',f(\bx_t',\bw_{D^i,t}))\|_2\big]\Big)\\
&\leq C_{p,\lambda}\mathbb{E}_{SGD}\big[\|\Delta\bw_{t}\|_2\big]
+ 2a_\ell a_\sigma C_{p,\lambda}\frac{\eta}{n} \sup_{\bx}\big\|\sum_{j\in N_\bx}e_{\cdot j}\bx_j\big\|_2  \\
&\quad +(a_\sigma^2+a_\ell) C_{p,\lambda}\eta\big(1-\frac{1}{n}\big)
 \big\|\sum_{j\in N_\bx}e_{\cdot j}\bx_j\big\|_2^2 \mathbb{E}_{SGD}\big[\|\Delta \bw_t\|_2\big]\\
&= C_{p,\lambda}\Big(1+(a_\sigma^2+a_\ell)\eta\big(1-\frac{1}{n}\big)\bg_{e}^2\Big)\mathbb{E}_{SGD}\big[\|\Delta\bw_{t}\|_2\big]
+\frac{2a_\ell a_\sigma \eta C_{p,\lambda}\bg_{e}}{n}.
\end{align*}

Finally, an application of the recursion  for the above inequality leads to   
\begin{align*}
\mathbb{E}_{SGD}\big[\|\Delta\bw_{T}\|_2\big] \leq \frac{2a_\ell a_\sigma \eta C_{p,\lambda}\bg_{e}}{n}\sum_{t=1}^T
\Big(C_{p,\lambda}\big(1+(a_\sigma^2+a_\ell)\eta\bg_{e}^2\big)\Big)^{t-1},
\end{align*}
where we use the fact that the parameter initialization is kept same for $D$ and $D^i$, e.g., $\bw_{D,0}=\bw_{D^i,0}$. Thus, the proof for Lemma \ref{recusweight} is completed.
\end{proof}

\section{Proof for Lemmas} \label{appendix:ProofLemmas}
\subsection{Deferred Proof for Lemma \ref{contrative}.}
% {\bf Proof for Lemma \ref{contrative}.}
\begin{proof}[Proof for Lemma \ref{contrative}.]
	Since the  regularization term $\|\bw\|_{\ell_p}^p$ is continuously differential and specially we get  $\partial(|w_j|^p)=p\cdot\hbox{sign}(w_j)\cdot|w_j|^{p-1}$ , where $\bw=(w_1,w_2,...,w_d)$. Note that all the terms   in \eqref{proxsolu}   are separable on $w_j$'s,  and
	%we just consider any given term  that is a function of $w_j$. 
	using the first-order condition for \eqref{proxsolu} leads to
	\begin{align}\label{firstoder}
	w_j^*-v_j+\lambda p\cdot\hbox{sign}(w_j^*)\cdot|w_j^*|^{p-1}=0,\quad j=1,2...,d.
	\end{align}
	
	If $w^*_j>0$, this follows from \eqref{firstoder} that
	$$
	w_j^*+\lambda p\cdot|w_j^*|^{p-1}=v_j,
	$$
	implying that
	$$
	0<w^*_j\leq \min\{v_j, (v_j/(\lambda p))^{1/(p-1)}\}.
	$$
	
	Otherwise, we obtain from \eqref{firstoder} that 
	$$
	-|w_j^*|-\lambda p\cdot|w_j^*|^{p-1}=v_j,
	$$
	implying that
	$$
	|w_j^*|\leq \min\{|v_j|, (|v_j|/(\lambda p))^{1/(p-1)}\}.
	$$
	
	Summing the previous two conclusions leads to our desired result. %$\Box$
\end{proof}

\subsection{Deferred Proof for Lemma \ref{convexdiff}.} \label{sub:convexdiff}
% {\bf Proof for Lemma \ref{convexdiff}.}
\begin{proof}[Proof for Lemma \ref{convexdiff}.]
	Let us introduce the notation:
	$$
	R_{D}(\bw):=\frac{1}{2}\|\bw-\bv_{D,i_t}\|_2^2,
	$$
	and similarly
	$$
	R_{D^i}(\bw):=\frac{1}{2}\|\bw-\bv_{D^i,i_t}\|_2^2.
	$$
	
	Recall that a convex function $g$ admits the following inequality:
	$$
	g(\bx+\theta(\bu-\bx))-g(\bx)\leq \theta(g(\bu)-g(\bx)),\quad \forall\,\bx,\bu.
	$$

	The convexity of  $R_{D}(\bw)$ and $R_{D^i}(\bw)$ immediately leads to
	\begin{align*}
	&R_{D^i}(\bw_{D,t+1}+\theta\Delta\bw_{t+1})-R_{D^i}(\bw_{D,t+1})\leq \theta \big(R_{D^i}(\bw_{D^i,t+1})-R_{D^i}(\bw_{D,t+1})\big).
	\end{align*}
	
	Then, switching the role of $\bw_{D,t+1}$ and $\bw_{D^i,t+1}$, we get
	\begin{align*}
	&R_{D^i}(\bw_{D^i,t+1}-\theta\Delta\bw_{t+1})-R_{D^i}(\bw_{D^i,t+1})\leq \theta \big(R_{D^i}(\bw_{D,t+1})-R_{D^i}(\bw_{D^i,t+1})\big).
	\end{align*}	
	
	Summing the two previous inequalities yields
	\begin{align}\label{sum1}
	R_{D^i}(\bw_{D,t+1}+\theta\Delta\bw_{t+1})-R_{D^i}(\bw_{D,t+1})+
	R_{D^i}(\bw_{D^i,t+1}-\theta\Delta\bw_{t+1})-R_{D^i}(\bw_{D^i,t+1})\leq 0.
	\end{align}
	
	Now, by the definitions of $\bw_{D,t+1}$ and $\bw_{D^i,t+1}$, we have
	\begin{align*}
	&R_{D}(\bw_{D,t+1})+\lambda_t\|\bw_{D,t+1}\|_{\ell_p}^p-\big[R_{D}(\bw_{D,t+1}+\theta\Delta\bw_{t+1})+
	\lambda_t\|\bw_{D,t+1}+\theta\Delta\bw_{t+1}\|_{\ell_p}^p\big]\leq 0,\\
	&R_{D^i}(\bw_{D^i,t+1})+\lambda_t\|\bw_{D^i,t+1}\|_{\ell_p}^p-\big[R_{D^i}(\bw_{D^i,t+1}-\theta\Delta\bw_{t+1})+
	\lambda_t\|\bw_{D^i,t+1}-\theta\Delta\bw_{t+1}\|_{\ell_p}^p\big]\leq 0.
	\end{align*}
	
	Then, summing the two previous inequalities and using \eqref{sum1} and \eqref{boundpara}, we obtain
	\begin{align}
	&\lambda_t\big(\|\bw_{D,t+1}\|_{\ell_p}^p-\|\bw_{D,t+1}+\theta\Delta\bw_{t+1}\|_{\ell_p}^p+
	\|\bw_{D^i,t+1}\|_{\ell_p}^p-\|\bw_{D^i,t+1}-\theta\Delta\bw_{t+1}\|_{\ell_p}^p\big)\nonumber\\
	&\leq R_{D}(\bw_{D,t+1}+\theta\Delta\bw_{t+1})-R_{D^i}(\bw_{D,t+1}+\theta\Delta\bw_{t+1})+
	R_{D^i}(\bw_{D,t+1})-R_{D}(\bw_{D,t+1})\nonumber\\
	&\leq \big(\|\bw_{D,t+1}+\theta\Delta\bw_{t+1}-\bv_{D,i_t}\|_2+\|\bw_{D,t+1}-\bv_{D^i,i_t}\|_2\big)
	\|\bv_{D,i_t}-\bv_{D^i,i_t}\|_2\nonumber\\
	&\leq \big(\|\bw_{D,t+1}+\theta\Delta\bw_{t+1}-\bv_{D,i_t}\|_2+
	\|\bw_{D,t+1}-\bv_{D^i,i_t}\|_2\big)\nonumber\\
	&\quad \times \big(\|\Delta\bw_{t}\|_2+\eta\|\nabla \ell(y,f(\bx,\bw_{D,t}))-\nabla \ell(y,f(\bx,\bw_{D^i,t}))\|_2 \big)\nonumber\\
	&\leq  7(B/\lambda)^{1/p}\big(\|\Delta\bw_{t}\|_2+\eta\|\nabla \ell(y,f(\bx,\bw_{D,t}))-\nabla \ell(y,f(\bx,\bw_{D^i,t}))\|_2 \big),
	\end{align}
	where we use the well-known property that  projection onto a convex set $\mathcal{C}$ is non-expansive, that is, for any two points $\bx,\bu$, $\|\Pi_{\mathcal{C}}(\bx)-\Pi_{\mathcal{C}}(\bu)\|_2\leq \|\bx-\bu\|_2$.
	This completes the proof of Lemma \ref{convexdiff}.  %\hspace*{13cm}$\Box$
\end{proof}

\subsection{Deferred Proof for Lemma \ref{derivativebound}.} \label{sub:derivativebound}
% {\bf Proof for Lemma \ref{derivativebound}.}
\begin{proof}[Proof for Lemma \ref{derivativebound}.]
	Recall that, the predictor function is defined as 
	$f(\bx,\bw)=\sigma\big(\sum_{j\in N_\bx}e_{\cdot j}\bx_j^T\bw\big)$. The gradient of $f(\bx,\bw)$ can be computed by
	$$
	\nabla f(\bx,\bw)=\sigma'\big(\sum_{j\in N_\bx}e_{\cdot j}\bx_j^T\bw\big) \sum_{j\in N_\bx}e_{\cdot j}\bx_j.
	$$
	%Since the loss function is Lipschitz continuous,  we get

	Using the chain-rule of compositional functions, we have
	\begin{align}\label{trnads}
	&\nabla \ell(y,f(\bx,\bw_{D,t}))-\nabla \ell(y,f(\bx,\bw_{D^i,t}))\nonumber\\
	&=\ell'(y,f(\bx,\bw_{D,t}))\nabla f(\bx,\bw_{D,t})-\ell'(y,f(\bx,\bw_{D^i,t}))\nabla f(\bx,\bw_{D^i,t})\nonumber\\
	&=\Big[\ell'(y,f(\bx,\bw_{D,t}))\sigma'\big(\sum_{j\in N_\bx}e_{\cdot j}\bx_j^T\bw_{D,t}\big) -
	\ell'(y,f(\bx,\bw_{D^i,t}))\sigma'\big(\sum_{j\in N_\bx}e_{\cdot j}\bx_j^T\bw_{D^i,t}\big)\Big] \times \Big(\sum_{j\in N_\bx}e_{\cdot j}\bx_j\Big).
	\end{align}
	
	Under Assumption 1, an application of the triangle inequality  leads to
	\begin{align*}
	&\Big|\ell'(y,f(\bx,\bw_{D,t}))\sigma'\big(\sum_{j\in N_\bx}e_{\cdot j}\bx_j^T\bw_{D,t}\big) -
	\ell'(y,f(\bx,\bw_{D^i,t}))\sigma'\big(\sum_{j\in N_\bx}e_{\cdot j}\bx_j^T\bw_{D^i,t}\big)\Big|\\
	&\leq \Big|\big[\ell'(y,f(\bx,\bw_{D,t})) -
	\ell'(y,f(\bx,\bw_{D^i,t})) \big] \sigma'\big(\sum_{j\in N_\bx}e_{\cdot j}\bx_j^T\bw_{D,t}\big)\Big|\\
	&\quad +\Big|\Big[\sigma'\big(\sum_{j\in N_\bx}e_{\cdot j}\bx_j^T\bw_{D,t}\big) -
	\sigma'\big(\sum_{j\in N_\bx}e_{\cdot j}\bx_j^T\bw_{D^i,t}\big)\Big] \ell'(y,f(\bx,\bw_{D^i,t}))\Big|\\
	&\leq a_\sigma \big|f(\bx,\bw_{D,t})-f(\bx,\bw_{D^i,t})\big|
	+ a_\ell \Big|\sum_{j\in N_\bx}e_{\cdot j}\bx_j^T(\bw_{D,t}-\bw_{D^i,t})\Big|\\
	&\leq (a_\sigma^2+a_\ell)\Big|\sum_{j\in N_\bx}e_{\cdot j}\bx_j^T(\bw_{D,t}-\bw_{D^i,t})\Big|\\
	&\leq (a_\sigma^2+a_\ell) \big\|\sum_{j\in N_\bx}e_{\cdot j}\bx_j\big\|_2 \|\Delta \bw_t\|_2.
	\end{align*}
	This together with \eqref{trnads} yields the desired result. %\hspace*{5cm}$\Box$
\end{proof}

\subsection{Deferred Proof for Lemma \ref{diffboundscase2}.} \label{sub:diffboundscase2}
% {\bf Proof for Lemma \ref{diffboundscase2}.}
\begin{proof}[Proof for Lemma \ref{diffboundscase2}.]
    Again using the chain-rule of compositional functions, we have
    \begin{align}\label{compderiva}
    &\nabla \ell(y,f(\bx,\bw_{D,t}))-\nabla \ell(y',f(\bx',\bw_{D^i,t}))\nonumber\\
    &=\ell'(y,f(\bx,\bw_{D,t}))\nabla f(\bx,\bw_{D,t})-\ell'(y',f(\bx',\bw_{D^i,t}))\nabla f(\bx',\bw_{D^i,t})\nonumber\\
    &=\ell'(y,f(\bx,\bw_{D,t}))\sigma'\big(\sum_{j\in N_\bx}e_{\cdot j}\bx_j^T\bw_{D,t}\big)\Big(\sum_{j\in N_\bx}e_{\cdot j}\bx_j\Big)
    -\ell'(y',f(\bx',\bw_{D^i,t}))\sigma'\big(\sum_{j\in N_{\bx'}}e_{\cdot j}(\bx')_j^T\bw_{D^i,t}\big)
     \Big(\sum_{j\in N_\bx'}e_{\cdot j}\bx'_j\Big)\nonumber\\
    &\leq 2a_\ell a_\sigma \sup_{\bx}\big\|\sum_{j\in N_\bx}e_{\cdot j}\bx_j\big\|_2,
    \end{align}
    where we use the fact from Assumption 1 that the first order derivatives of $\ell$ and $\sigma$ are both bounded. This completes the proof of Lemma \ref{diffboundscase2}. %\hspace*{6cm}$\Box$
\end{proof}

\subsection{Additional Lemma}
\begin{lem}\label{eigenbound}
	Let $G=(V,E)$ be an undirected graph with a weighted adjacency matrix 
	$g(\bL)$ and $\lambda^{max}_{G}$ be the maximum absolute eigenvalue of $g(\bL)$. Let $G_\bx$
	be the ego-graph of a node $\bx \in V$ with corresponding maximum absolute eigenvalue 
	$\lambda^{max}_{G_\bx}$. Then the following eigenvalue bound holds for all $\bx$,
	$$
	\lambda^{max}_{G_\bx}\leq \lambda^{max}_{G}.
	$$
\end{lem}
Lemma \ref{eigenbound} is proved in reference \cite{verma2019stability}, where they also showed that
$\bg_{e}$ can be upper bounded in term of the largest absolute eigenvalue of $g_\bx(\bL)$. That is,
$\bg_{e}\leq \lambda^{max}_{G_\bx}$ for all $\bx$. This  follows from Lemma \ref{eigenbound} that 
\begin{align}\label{eigexp}
\bg_{e}\leq \lambda^{max}_{G}.
\end{align}

Finally, plugging  \eqref{eigexp} and Lemma \ref{recusweight} into \eqref{reduce} yields the following result:
\begin{align}
 \beta_n\leq a_{\ell}^2a_{\sigma}^2\lambda^{max}_{G}  \frac{ \eta C_{p,\lambda}\bg_{e}}{n}\sum_{t=1}^T\Big(C_{p,\lambda}\big(1+(a_\sigma^2+a_\ell)\eta\bg_{e}^2\big)\Big)^{t-1}.
\end{align}

\section{Additional Experiments and Setup Details} \label{app:Experiments}
\subsection{General Setup}
We conduct experiments on three citation network datasets: Citeseer, Cora, and Pubmed \cite{sen2008collective}. In every datasets, each document represents as spare bag-of-words feature vector, and the relationship between documents consists of a list of citation links, which can be treated as undirected edges and help to construct the adjacency matrix. These documents can be divided into different classes and have the corresponding class label. The data statistics for the datasets used in Section 5 are summarized in \cref{tab:data_statis}. 

\begin{table}[htp]
    \centering
    \begin{tabular}{l|rrrrr}
        \toprule
         \textbf{Dataset}   &\textbf{Nodes} &\textbf{Edges}   &\textbf{Classes}   &\textbf{Features} \\
         %& \textsc{Label rates}\\
         \midrule
         Citeseer   &  $3,327$     & $4,732$     & $6$ & $3,703$ \\ %& $0.036$\\
         Cora       &  $2,708$     & $5,429$     & $7$ & $1,433$ \\ %& $0.052$\\
         Pubmed     &  $19,717$    & $44,338$    & $3$ & $500$   \\ %& $0.003$\\
        \bottomrule
    \end{tabular}
    \caption{Dataset Statistics.}
    \label{tab:data_statis}
\end{table}

\subsection{Additional Experimental Results}
In this section, we investigate the stability and generalization of $\ell_p$-regularized stochastic learning for GCN under different normalization steps. We mainly consider the implication of our results in following designing graph convolution filters:

\begin{enumerate} %[leftmargin=*]
	\item[(I)] Unnormalized Graph Filters: $g(\bL)=\bA+\mathbf{I}$;
    \item[(II)] Normalized Graph Filters: $g(\bL)=\mathbf{D}^{-1/2}\mathbf{A}\mathbf{D}^{-1/2}+\mathbf{I}$;
    \item[(III)] Random Walk Graph Filters: $g(\bL)=\mathbf{D}^{-1}\mathbf{A}+\mathbf{I}$;
    \item[(IV)] Augmented Normalized Graph Filters: $g(\bL)=(\mathbf{D} + \mathbf{I})^{-1/2} ( \mathbf{A} + \mathbf{I} ) (\mathbf{D} + \mathbf{I})^{-1/2}$.
\end{enumerate}

For each experiment, we initialize the parameters of GCN models with the same random seeds and then train all models for a maximum of 200 epochs using the proposed Inexact Proximal SGD. We repeat the experiments ten times and report the average performance and the standard variance. 
We quantitatively measure the generalization gap defined as the absolute difference between the training and test errors and the parameter distance $\sqrt{\nicefrac{\|\widehat{\mathbf{w}}-\widehat{\mathbf{w}}^{\prime}\|^{2}}{\big(\|\widehat{\mathbf{w}}\|^{2}+\|\widehat{\mathbf{w}}^{\prime}\|^{2}\big)}}$ between the parameters $\widehat{\mathbf{w}}$ and $\widehat{\mathbf{w}}^{\prime}$ of two models trained on two copies of the data differing in a random substitution. 

\begin{figure*}[!h]
	\centering
	\includegraphics[width=0.95\linewidth]{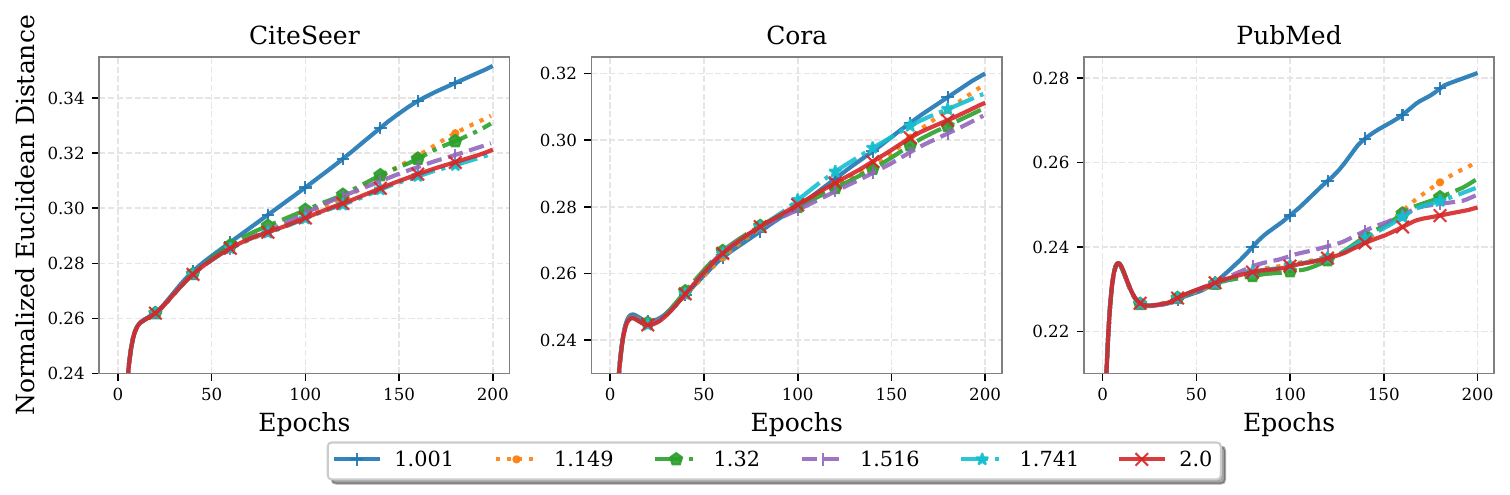}
	\caption{\textbf{Normalized euclidean distance between parameters of two models trained under various $p$ on three datasets.}}
	\label{fig:parameter_diff}
\end{figure*}

\begin{figure*}[!h]
	\centering	
	\includegraphics[width=0.95\linewidth]{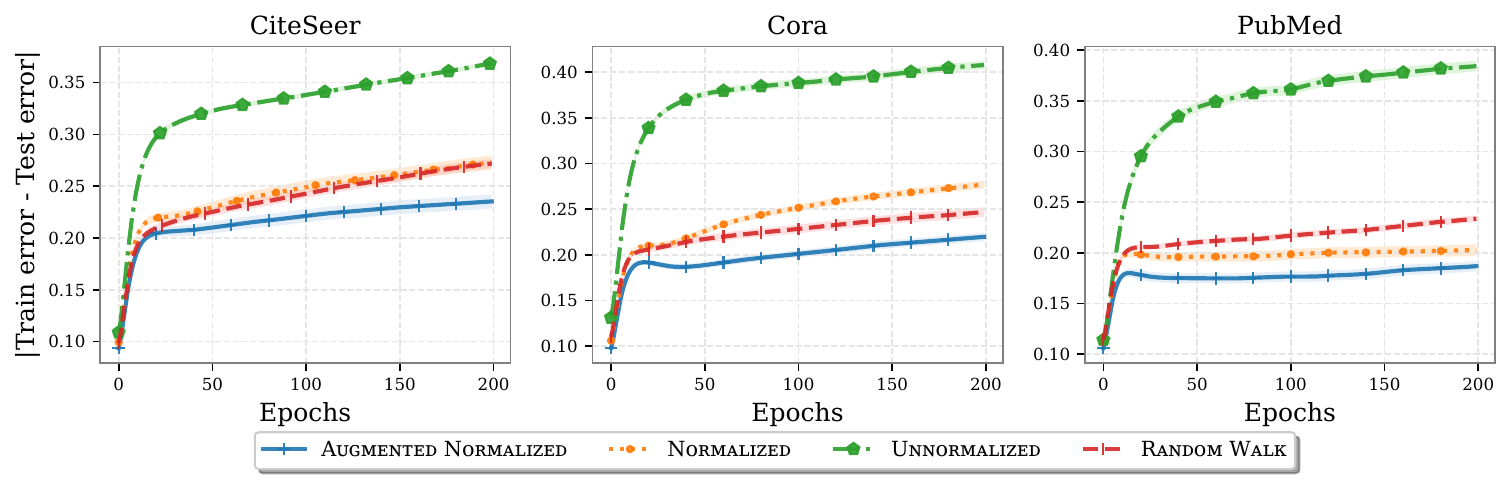}
	\caption{\textbf{Generalization Gaps under different normalization steps on three datasets.}}
	\label{fig:4}
\end{figure*}

The unnormalized graph convolution filters and Random Walk Graph Filters show a significantly higher generalization gap than the normalized ones. The results hold consistently across the three datasets under different $p$. Hence, these empirical results are also consistent with our generalization error bound. We note that the generalization gap becomes constant after a certain number of iterations.

\clearpage % don't remove
\end{appendices}

\end{document}